\pdfoutput=1
\documentclass[sigconf]{acmart}

\AtBeginDocument{%
  \providecommand\BibTeX{{%
    \normalfont B\kern-0.5em{\scshape i\kern-0.25em b}\kern-0.8em\TeX}}}

\usepackage{balance}
\usepackage{hyperref}
\usepackage{booktabs}
\usepackage{multirow}
\usepackage{siunitx}
\usepackage{amsmath,amsfonts}
\usepackage{algorithm}
\usepackage{graphicx}
\usepackage{textcomp}
\usepackage{url}
\usepackage{tabularray}
\usepackage{balance}
\usepackage{graphics, subcaption, algorithm, amsmath, hyperref, multirow, float}
\usepackage[noend]{algpseudocode}
\usepackage[super]{nth}
\usepackage[table]{xcolor}
\graphicspath{{figures/}}
\usepackage{comment}

\copyrightyear{2026}
\acmYear{2026}
\setcopyright{cc}
\setcctype{by}
\acmConference[GECCO Companion '26]{Genetic and Evolutionary Computation Conference}{July 13--17, 2026}{San Jose, Costa Rica}
\acmBooktitle{Genetic and Evolutionary Computation Conference (GECCO Companion '26), July 13--17, 2026, San Jose, Costa Rica}
\acmDOI{10.1145/3795101.3805275}
\acmISBN{979-8-4007-2488-6/2026/07}

\begin{document}

\title{Big Brains and Changing Environments: Cause or Consequence?}

\author{Sian Heesom-Green, Jonathan Shock, Geoff Nitschke}
 \email{HSMSIA001@myuct.ac.za, jonathan.shock@uct.ac.za, gnitschke@cs.uct.za}
 \affiliation{
  \institution{Department of Computer Science, University of Cape Town}
  \city{Cape Town}
  \country{South Africa}
 }
 

\begin{abstract}
Large brains are metabolically costly, and associations with changing environments do not imply they evolved there, as the \textit{Cognitive Buffer Hypothesis} (CBH) would suggest. They may instead evolve in stable conditions and later facilitate colonization of changing environments. Using neuro-evolution in an artificial seasonal foraging task, we compared agents evolving exclusively in changing environments to agents first evolved in static environments before transitioning. Results show that larger neural networks in dynamic environments arise mainly from prior static evolution, achieving superior performance under unpredictable changes. Our results challenge strict CBH predictions, provide agent-based (computational) support for a colonization-based account and highlight the role of evolutionary history in brain size evolution.

\end{abstract}



\maketitle

\section{Introduction}\label{sec:intro}
Associations between large brains and changing environments have been identified in birds \cite{vincze2016light, sayol2016environmental, schuck2008cognition}, but do these associations imply causality, as suggested by the Cognitive Buffer Hypothesis (CBH)? CBH posits that larger brains evolved in response to changing environments by enabling enhanced behavioral flexibility and learning, helping buffer the effects of seasonality \cite{allman1993brain,sol2009revisiting,michaud2022impact}.

However, large brains are metabolically costly, and energy constraints may strongly shape their evolution. The Expensive Brain Hypothesis (EBH) frames this cost as a fundamental bottleneck: increases in brain size are only sustainable when energy budgets allow, either through greater energy acquisition or the reallocation of energy from other vital organs \cite{isler2009expensive}. In conditions where energy intake is more difficult, such as in changing or unpredictable environments, these constraints may limit rather than promote larger brains. Consistent with this, both biological and artificial systems show that smaller, more energy-efficient brains can be favored under variable conditions \cite{heesom2025energy,van2010effects,luo2017seasonality,graber2017social,weisbecker2015evolution,van2012large}.

If environmental variability does not consistently favor larger brains, what explains their observed association in birds today? One alternative to CBH is that large brains facilitate the colonization of variable environments rather than evolving within them \cite{fristoe2017big,sayol2016environmental,van2011influence}. \citet{fristoe2017big} showed that birds with larger relative brain sizes were more resilient to environmental variation, yet increases in brain size typically preceded, rather than followed, expansion into more variable habitats. This suggests that large brains may evolve under stable, energy-permissive conditions and, as a consequence, facilitate expansion into more dynamic environments.

Neuro-evolution, the artificial evolution of Artificial Neural Networks (ANNs), provides a controlled framework for simulating evolutionary dynamics across diverse environmental and ecological scenarios and trajectories \cite{miikkulainen2025neuroevolution}. In this study, we employ neuro-evolution in an artificial foraging task with both predictable and unpredictable seasonal variation, incorporating explicit energy costs on ANN size. We compare agents that evolve initially in static environments before transitioning to changing environments with those evolving solely in changing environments, asking whether prior evolution in static environments facilitates the emergence of larger ANNs and improved performance under environmental variability. By incorporating evolutionary trajectory, this framework enables a more causally grounded interpretation of the relationship between brain size and environmental variability.

\section{Methods and Experiments}\label{sec:methods}
This section describes the task environment (section \ref{section:task-environment}), neuro-evolution process (section  \ref{section:neuro-evolution}), agent lifetime learning (section \ref{section:lifetime-learning}), and experiments (section \ref{sec:exps}).

\begin{figure}[t]
    \centering
    \includegraphics[width=0.38\textwidth]{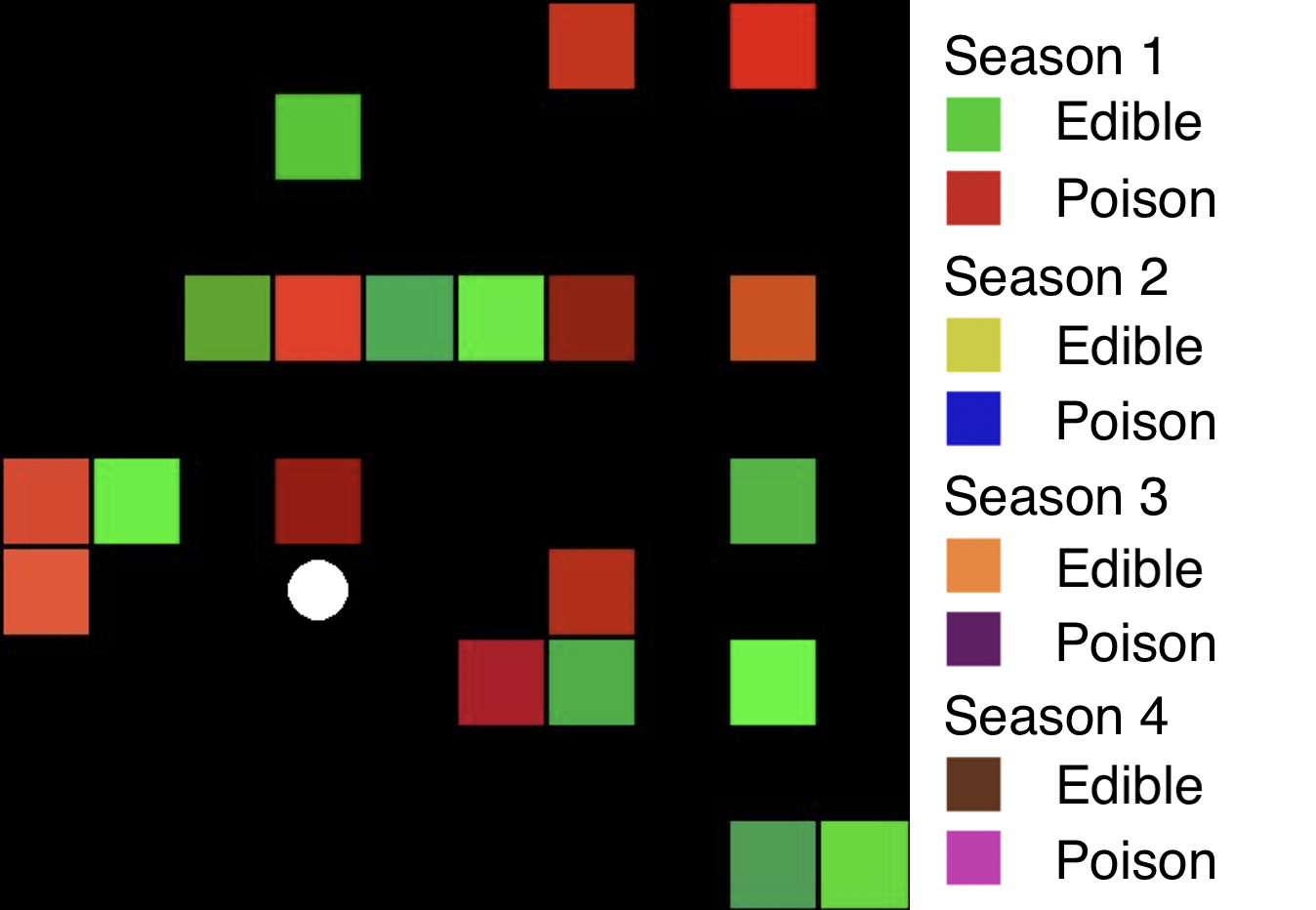}
    \caption{10×10 grid-world with cell colors denoting food type. The agent is represented by a white circle. The legend provides an example of seasonal changes in food color.
    }
    \label{fig:env}
\end{figure}

\subsection{Task Environment}
Agents forage for edible foods while avoiding poisonous items in a 10×10 grid-world, with cell color indicating food type (Figure \ref{fig:env}) \citep{heesom2025energy}. They can move up, down, left, right, or eat, and observe a 7×7 window of nearby RGB cell colors (147 values), their current location (1 value), and previous reward (1 value). Each 100-step episode begins with 10 edible and 10 poisonous items randomly placed in the grid; consumed items are immediately replaced. Base RGB colors for each food type are randomly initialized, with individual items varying slightly ($\pm$0.2 per channel).

\paragraph{Dynamic Environments} In dynamic environments, the color assigned to each food type changes across seasons (example shown in Figure \ref{fig:env} legend). Seasons follow a fixed sequence with equal duration.  
In predictable environments, food color mappings are fixed within each season, whereas in unpredictable environments, they are regenerated at every seasonal change.

\paragraph{Rewards / energy dynamics} Agents expend energy (Equation \ref{eq:E}) per time-step, gain energy (+1) from edible foods, and lose energy (-1) from poisonous foods. Energy expenditure scales with ANN size to reflect the metabolic cost of larger brains:
\begin{equation}
E = 0.01 \times \frac{N_S^{\text{current}}}{N_S^{\text{gen$_0$}}}
\label{eq:E}
\end{equation}
Where, $N_S^{\text{current}}$ is the current ANN size and $N_S^{\text{gen$_0$}}$ is the initial size (generation 0)\footnote{$E$=$0.01$ per time-step initially so total energy expenditure over an episode sums to 1.}. ANN size ($N_S$, Equation \ref{eq:Ns}) is defined as the total number of free parameters \citep{nagar2019cost,nagar2019costBodies,heesom2025energy}:
\begin{equation}
N_S = (\text{non-input nodes}) + (\text{connections})
\label{eq:Ns}
\end{equation}
\label{section:task-environment}

\subsection{Agent Neuro-evolution}\label{section:neuro-evolution}

Agent ANNs evolve using the standard NEAT procedure \cite{stanley2002evolving}. We evolve a population of 150 genomes (ANNs) over 1100 generations, each initialized with a partially connected ANN (input-output connections occur with 50\% probability), with no hidden nodes and randomly initialized weights, biases, and per-connection learning parameters\footnote{Full NEAT evolutionary parameter details \href{https://github.com/sianmay/CBH-versus-colonization/blob/main/neat_config}{here}.}.

Evolution follows standard NEAT mechanisms, including speciation, selection, reproduction, and replacement. Within each non-stagnant species (no improvement over 15 generations), fittest individuals are selected via rank-based selection to generate offspring through crossover and mutation. The next generation is formed from these offspring, with elitism applied by retaining the fittest genome in species containing more than five individuals.

Fitness is evaluated over 10 independent trials with different seeds and food color configurations. Each trial consists of 100 episodes in which agents update their ANN weights online\footnote{Weight updates are not carried over between trials or across generations.} via Hebbian learning (see section \ref{section:lifetime-learning}). Episodes 1-50 are used for exploration and 51-100 for computing trial fitness as the mean accumulated reward. Final fitness is averaged across trials. Multiple short trials reduce overfitting to specific episode sequences or colors, with seeds re-sampled every 50 generations during evolution.

\subsection{Lifetime Learning via Hebbian Learning}
\label{section:lifetime-learning}
Within each generation, agents adapt online via reward-modulated Hebbian learning \cite{gerstner2014neuronal,pfeiffer2010reward}. Weight updates follow Hebbian rules
(Equations 3-5):
\begin{equation}
H(n_i, n_j) = n_i \cdot n_j
\end{equation}
where $n_i$ and $n_j$ are presynaptic and postsynaptic activations. Temporal credit assignment is implemented via eligibility traces:
\begin{equation}
e_{ij} \leftarrow \tau_{ij} e_{ij} + H(n_i, n_j)
\end{equation}
with decay $\tau_{ij}$. Reward modulation adjusts weights according to:
\begin{equation}
w_{ij} \leftarrow w_{ij} + R , \eta_{ij} , e_{ij}
\end{equation}
where $R$ is the instantaneous reward minus its running average, and $\eta_{ij}$ is the connection-specific learning rate \cite{gerstner2014neuronal}.

\begin{table}[t]
\caption{Environment transitions over evolutionary time 
}
\begin{tabular}{|l|llll|}
\hline
\rowcolor[HTML]{C0C0C0} 
\textbf{\begin{tabular}[c]{@{}l@{}}Transition\\ Type\end{tabular}} & \multicolumn{1}{l|}{\cellcolor[HTML]{C0C0C0}\textbf{\begin{tabular}[c]{@{}l@{}}Gen \\ 0-99\end{tabular}}} & \multicolumn{1}{l|}{\cellcolor[HTML]{C0C0C0}\textbf{\begin{tabular}[c]{@{}l@{}}Gen \\ 100-199\end{tabular}}} & \multicolumn{1}{l|}{\cellcolor[HTML]{C0C0C0}\textbf{\begin{tabular}[c]{@{}l@{}}Gen \\ 200-299\end{tabular}}} & \textbf{\begin{tabular}[c]{@{}l@{}}Gen \\ 300+\end{tabular}} \\ \hline
None & \multicolumn{4}{l|}{4 seasons} \\ \hline
Sudden & \multicolumn{2}{l|}{1 season} & \multicolumn{2}{l|}{4 seasons} \\ \hline
Gradual & \multicolumn{1}{l|}{1 season} & \multicolumn{1}{l|}{2 seasons} & \multicolumn{1}{l|}{3 seasons} & 4 seasons \\ \hline
\end{tabular}
\label{tab:transitions}
\end{table}

\begin{table}[t]
\caption{Experimental conditions and evaluation}
\begin{tabular}{|l|l|l|}
\hline
\rowcolor[HTML]{C0C0C0} 
\multicolumn{1}{|c|}{\cellcolor[HTML]{C0C0C0}\textbf{\begin{tabular}[c]{@{}c@{}}Seasonal\\ Changes\end{tabular}}} & \multicolumn{1}{c|}{\cellcolor[HTML]{C0C0C0}\textbf{\begin{tabular}[c]{@{}c@{}}Environment\\ Transition Type\end{tabular}}} & \multicolumn{1}{c|}{\cellcolor[HTML]{C0C0C0}\textbf{\begin{tabular}[c]{@{}c@{}}Evaluation\\ Metrics\end{tabular}}} \\ \hline
 & None &  \\ \cline{2-2}
 & Sudden &  \\ \cline{2-2}
\multirow{-3}{*}{Predictable} & Gradual & \multirow{-3}{*}{\begin{tabular}[c]{@{}l@{}}ANN size (Ns),\\ Task Performance\end{tabular}} \\ \hline
 & None &  \\ \cline{2-2}
 & Sudden &  \\ \cline{2-2}
\multirow{-3}{*}{Unpredictable} & Gradual & \multirow{-3}{*}{\begin{tabular}[c]{@{}l@{}}ANN size (Ns),\\ Task Performance\end{tabular}} \\ \hline
\end{tabular}
\label{tab:exp-sets}
\end{table}

\subsection{Experiments}\label{sec:exps}
Experiments\footnote{Source code: \url{https://github.com/sianmay/CBH-versus-colonization}} investigate whether large ANNs emerge primarily under static, energy-permissive conditions and subsequently facilitate adaptation to changing environments, or whether they evolve directly in response to environmental variability. To test this, we compare agents evolved exclusively in changing environments with agents that first undergo evolution in static environments before being transitioned to changing environments, either abruptly or gradually (Table \ref{tab:transitions}). Each experimental condition (Table \ref{tab:exp-sets}) is evaluated over 20 independent evolutionary runs. Across all runs, we record ANN size ($N_S$, Equation \ref{eq:Ns}) and task performance, defined as net energy intake (edible minus poisonous food consumption). This metric differs from the fitness function in that it excludes energy expenditure, enabling fair comparisons across agents with varying ANN sizes \citep{heesom2025energy}.

\begin{figure*}[t]
\centering
\includegraphics[width=0.40\textwidth]{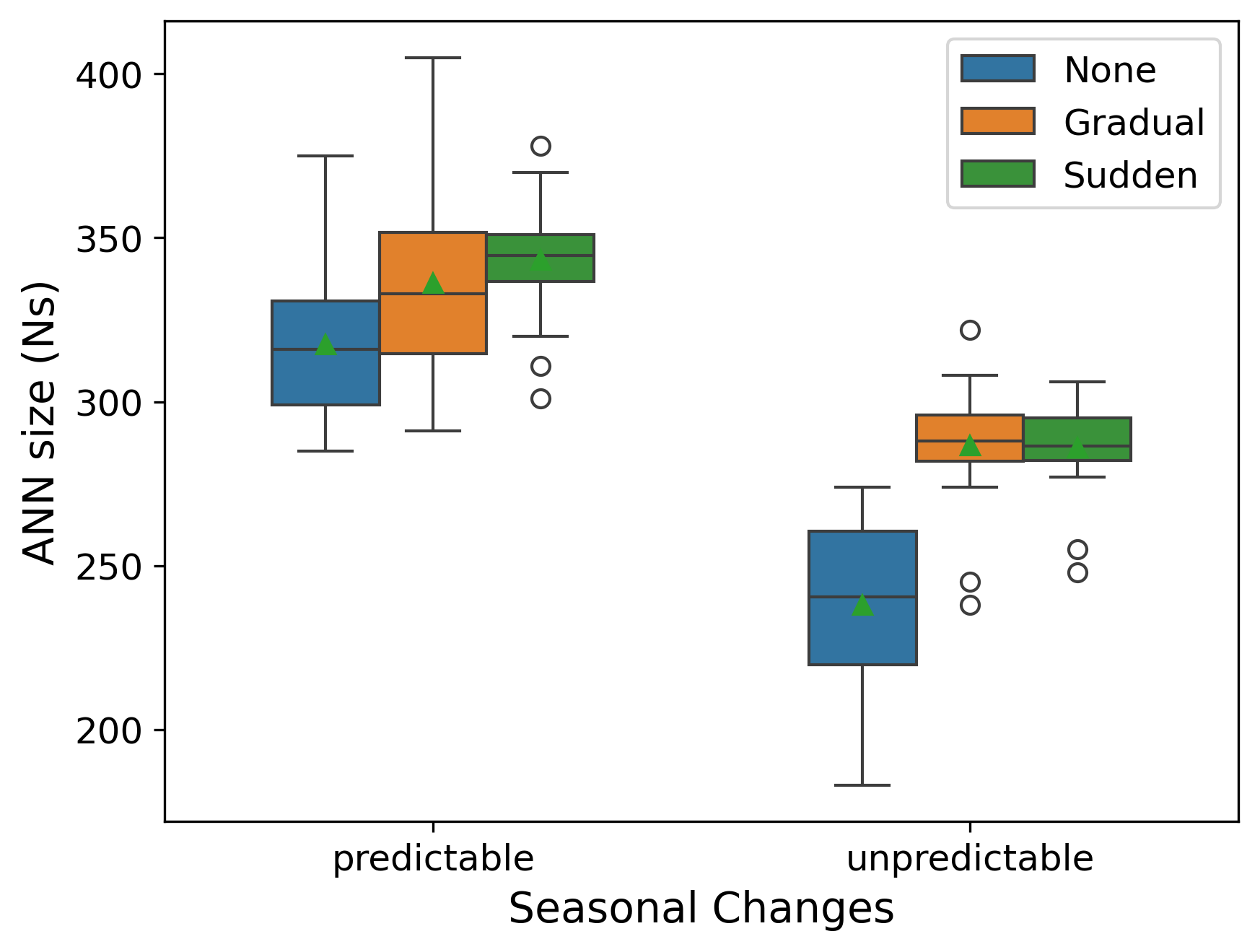}
\includegraphics[width=0.40\textwidth]{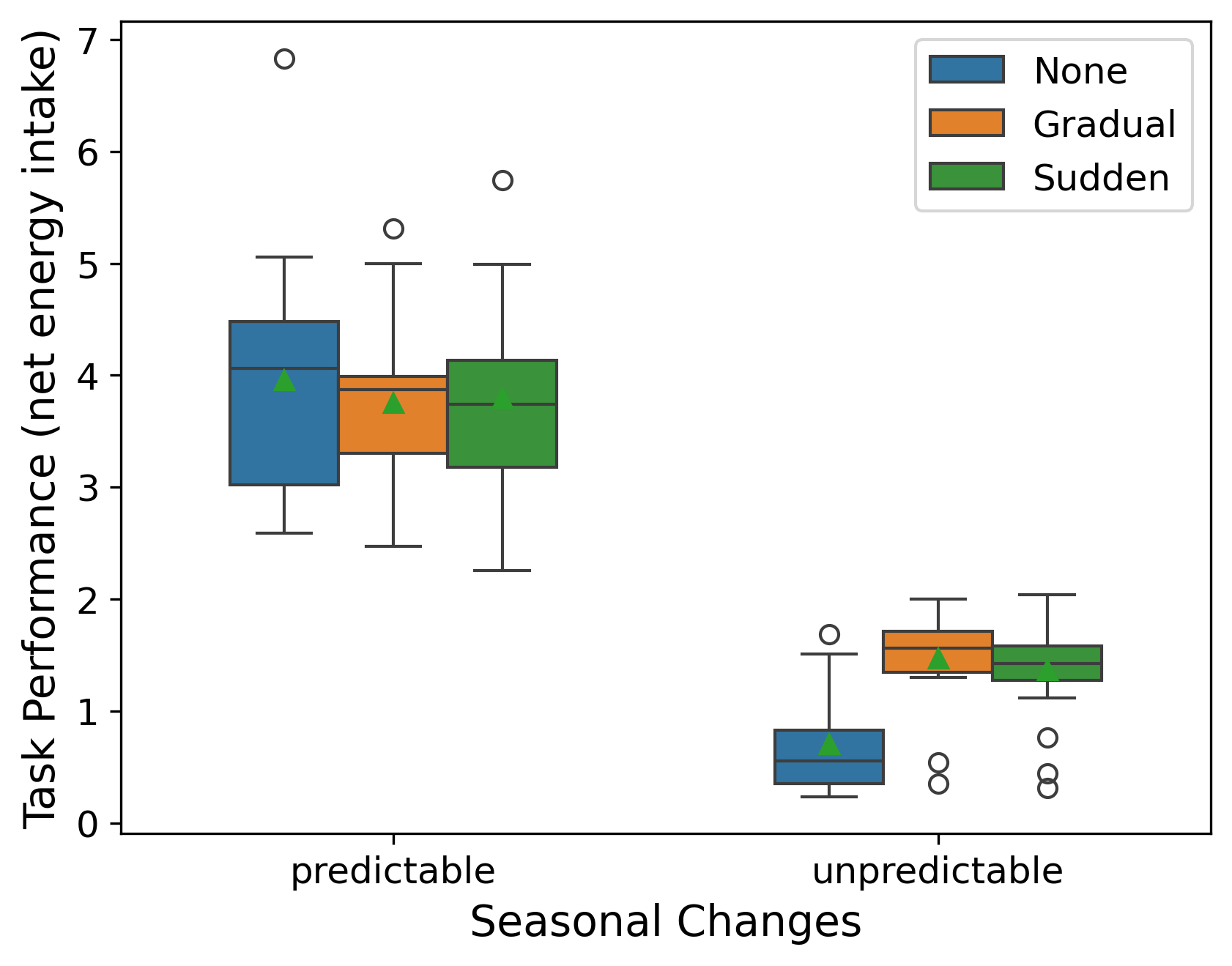}
\includegraphics[width=0.40\textwidth]{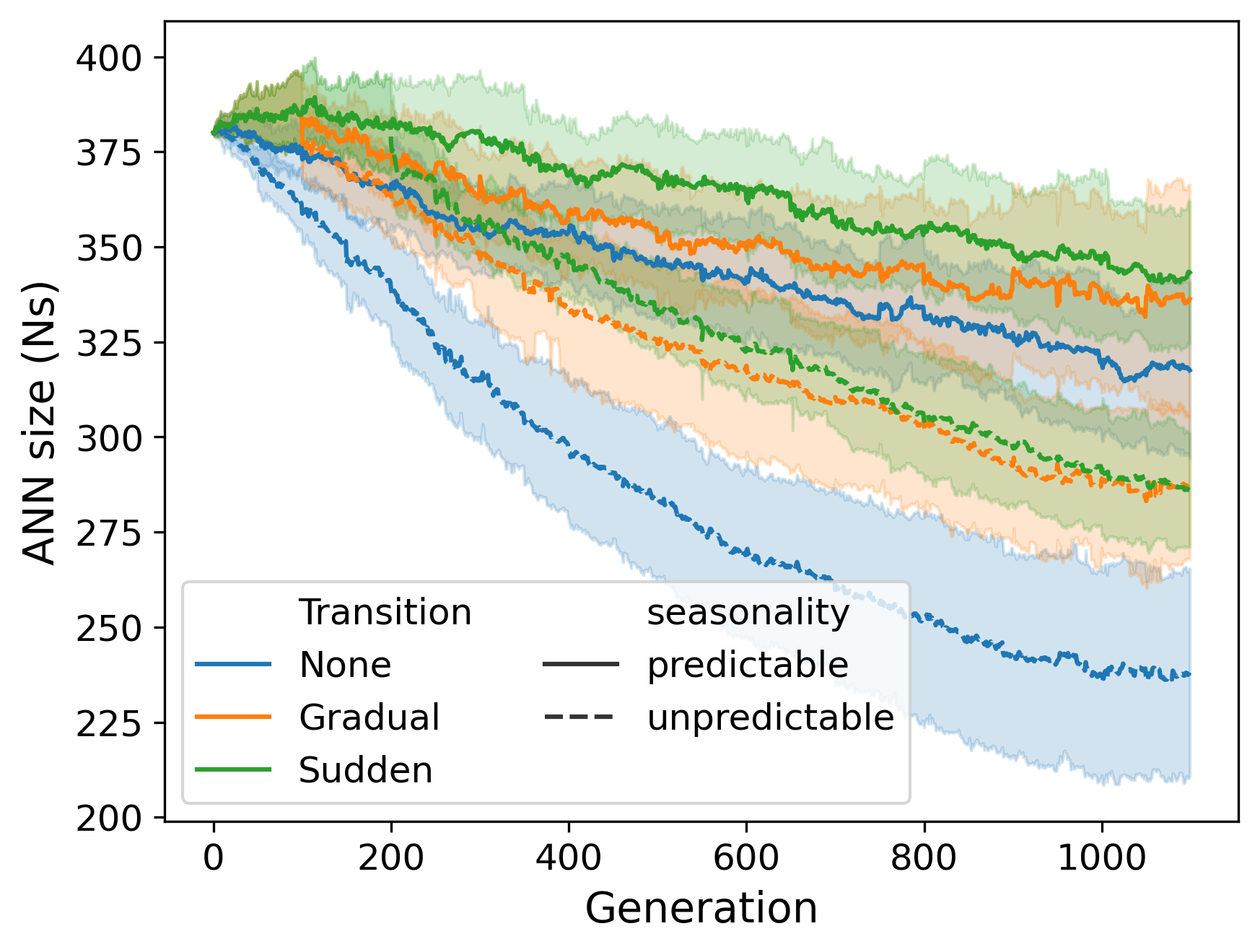}
\includegraphics[width=0.40\textwidth]{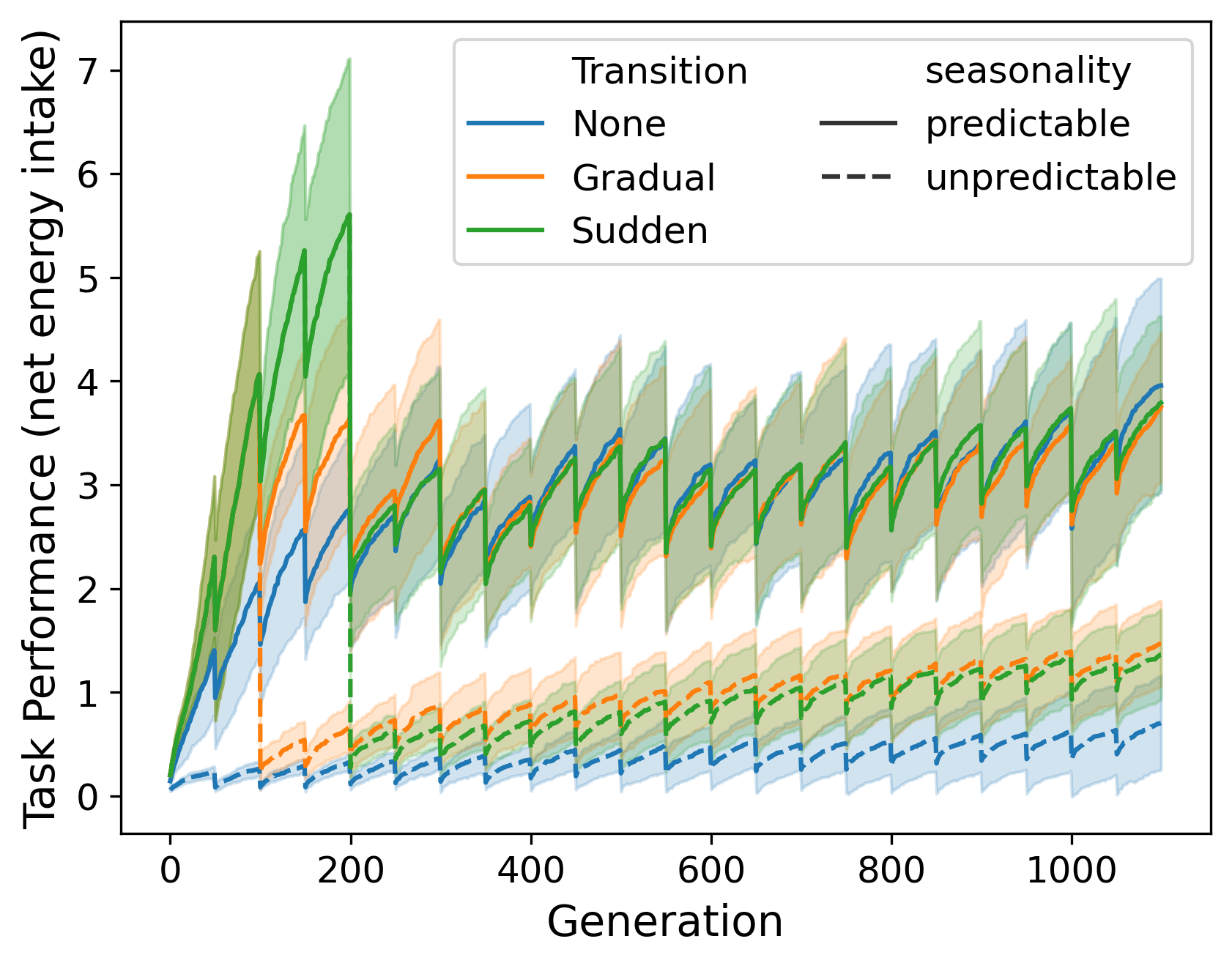}
\caption{ANN size ($N_S$, left panels) and task performance (right panels) for the fittest genome (averaged over 20 runs), across seasonal change regimes (predictable and unpredictable) and evolutionary transition scenarios (described in Table \ref{tab:transitions}). Box plots (top) show final evolved $N_S$ and task performance at the end of evolution per scenario. Line plots (bottom) show evolutionary trajectories of $N_S$ (left) and task performance (right) of the current fittest genome across generations for each scenario.
}\label{fig:plots}
\end{figure*}

\begin{table}[t]
\centering
\caption{Dunn’s post hoc tests (Bonferroni-corrected) for pairwise comparisons of ANN size and task performance across transition scenarios. Significant results ($p < 0.05$) are in bold.}
\setlength{\tabcolsep}{4pt}
\begin{tabular}{|p{2cm}|p{2.87cm}|p{2.87cm}|}
\hline
\rowcolor[HTML]{C0C0C0} 
\textbf{Seasonal Changes} & \textbf{ANN size} & \textbf{Task Performance} \\ \hline
 & \textbf{Sudden $>$ None} & Sudden $==$ None \\ \cline{2-3} 
 & Gradual $==$ None & Gradual $==$ None \\ \cline{2-3} 
\multirow{-3}{*}{Predictable} & Sudden $==$ Gradual & Sudden $==$ Gradual \\ \hline
 & \textbf{Sudden $>$ None} & \textbf{Sudden $>$ None} \\ \cline{2-3} 
 & \textbf{Gradual $>$ None} & \textbf{Gradual $>$ None} \\ \cline{2-3} 
\multirow{-3}{*}{Unpredictable} & Sudden $==$ Gradual & Sudden $==$ Gradual \\ \hline
\end{tabular}
\label{tab:dunn}
\end{table}

\section{Results \& Discussion}\label{sec:results}
Figure \ref{fig:plots} presents ANN size and task performance of the fittest genomes across evolutionary transition scenarios under both predictable and unpredictable seasonal conditions. Table \ref{tab:dunn} reports pairwise comparisons of ANN size and task performance across transition scenarios using Dunn’s post hoc tests (with Bonferroni correction) following Kruskal-Wallis (KW) tests \cite{kruskal1952use,dunn1964multiple}.

Early evolutionary conditions strongly influenced ANN size ($N_S$). Selection favored larger ANNs in static environments, where energy acquisition is easier, than in predictable and unpredictable changing environments (Spearman rank correlation for $N_S$ over generations before transitions: Static: $\rho = 0.037$, changing environments: $\rho < 0$, all $p < 0.05$), consistent with the Expensive Brain Hypothesis (EBH) \cite{isler2009expensive}. Although ANN sizes declined following transition to changing environments, agents with prior static evolution retained significantly larger networks at the end of the evolutionary run (KW test, $p<0.05$; Table \ref{tab:dunn}), indicating a lasting influence of early evolutionary history.

Under predictable changes, no significant differences in task performance were observed between transition scenarios (KW test, $p \geq 0.05$). Similarly, field observations report comparable foraging efficiency between larger-brained primates and smaller-brained procyonids, and empirical studies have reported both specialists and generalists under predictable seasonal variation, despite differences in brain size \citep{mettke2014cognitive,buchi2014coexistence,hirsch2024smarter}.

Under unpredictable changes, agents with prior static evolution performed significantly better than those evolving exclusively within changing environments (KW test, $p < 0.05$; Table \ref{tab:dunn}), consistent with evidence that larger brains enhance survival and colonization in harsh or unpredictable conditions \citep{vincze2016light,wagnon2020smaller,sol2000behavioural,sol2005big,sol2008brain,amiel2011smart}.

Associations between large ANNs and changing environments in our simulations therefore primarily reflect prior evolution under static conditions, which later confers advantages in variable environments, particularly when changes are unpredictable. These results align more closely with a colonization-based account than with the CBH, in which large brains precede and facilitate the colonization of variable habitats rather than evolving in response to them \cite{fristoe2017big}. This underscores the importance of investigating evolutionary trajectories, highlighting that associations do not necessarily imply causality \cite{dunbar2017there,heldstab2022economics,gould1982exaptation,fristoe2017big}.

\section{Conclusions}
This study examined whether associations between large brains and changing environments reflect selection within such environments, or instead arise from prior evolution under more stable conditions, subsequently facilitating colonization of variable environments. Using artificial neuro-evolution with explicit energy constraints, we show that changing environments do not reliably select for larger neural architectures. Instead, larger ANNs primarily arise from prior evolution in static environments, where energy acquisition is easier, with these agents achieving comparable performance under predictable changes and superior performance under unpredictable changes relative to those evolving exclusively in changing environments. More broadly, this work shows that correlations between brain size and changing environments do not necessarily imply causality, underscoring the need to account for evolutionary history when interpreting such associations within evolutionary theory. While artificial agents allow precise control over environmental dynamics and evolutionary history, the simplified setup limits generalization to biological systems. Ongoing work is exploring richer environmental variability, energy costs (for example, corresponding to variable agent morphologies \cite{mailer2021evolving, WatsonNitschke2015}), and alternative initialization and encoding schemes.


\section{Acknowledgments}
Compute was performed using the University of Cape Town’s ICTS High Performance Computing cluster: \url{hpc.uct.ac.za}


\bibliographystyle{ACM-Reference-Format}
\balance 
\bibliography{references}


\begin{thebibliography}{36}


\ifx \showCODEN    \undefined \def \showCODEN     #1{\unskip}     \fi
\ifx \showDOI      \undefined \def \showDOI       #1{#1}\fi
\ifx \showISBNx    \undefined \def \showISBNx     #1{\unskip}     \fi
\ifx \showISBNxiii \undefined \def \showISBNxiii  #1{\unskip}     \fi
\ifx \showISSN     \undefined \def \showISSN      #1{\unskip}     \fi
\ifx \showLCCN     \undefined \def \showLCCN      #1{\unskip}     \fi
\ifx \shownote     \undefined \def \shownote      #1{#1}          \fi
\ifx \showarticletitle \undefined \def \showarticletitle #1{#1}   \fi
\ifx \showURL      \undefined \def \showURL       {\relax}        \fi
\providecommand\bibfield[2]{#2}
\providecommand\bibinfo[2]{#2}
\providecommand\natexlab[1]{#1}
\providecommand\showeprint[2][]{arXiv:#2}

\bibitem[\protect\citeauthoryear{Allman, McLaughlin, and Hakeem}{Allman
  et~al\mbox{.}}{1993}]%
        {allman1993brain}
\bibfield{author}{\bibinfo{person}{John Allman}, \bibinfo{person}{Todd
  McLaughlin}, {and} \bibinfo{person}{Atiya Hakeem}.}
  \bibinfo{year}{1993}\natexlab{}.
\newblock \showarticletitle{Brain weight and life-span in primate species.}
\newblock \bibinfo{journal}{\emph{Proceedings of the National Academy of
  Sciences}} \bibinfo{volume}{90}, \bibinfo{number}{1} (\bibinfo{year}{1993}),
  \bibinfo{pages}{118--122}.
\newblock


\bibitem[\protect\citeauthoryear{Amiel, Tingley, and Shine}{Amiel
  et~al\mbox{.}}{2011}]%
        {amiel2011smart}
\bibfield{author}{\bibinfo{person}{Joshua~J Amiel}, \bibinfo{person}{Reid
  Tingley}, {and} \bibinfo{person}{Richard Shine}.}
  \bibinfo{year}{2011}\natexlab{}.
\newblock \showarticletitle{Smart moves: effects of relative brain size on
  establishment success of invasive amphibians and reptiles}.
\newblock \bibinfo{journal}{\emph{PLoS One}} \bibinfo{volume}{6},
  \bibinfo{number}{4} (\bibinfo{year}{2011}), \bibinfo{pages}{e18277}.
\newblock


\bibitem[\protect\citeauthoryear{B{\"u}chi and Vuilleumier}{B{\"u}chi and
  Vuilleumier}{2014}]%
        {buchi2014coexistence}
\bibfield{author}{\bibinfo{person}{Lucie B{\"u}chi} {and}
  \bibinfo{person}{S{\'e}verine Vuilleumier}.} \bibinfo{year}{2014}\natexlab{}.
\newblock \showarticletitle{Coexistence of specialist and generalist species is
  shaped by dispersal and environmental factors}.
\newblock \bibinfo{journal}{\emph{The American Naturalist}}
  \bibinfo{volume}{183}, \bibinfo{number}{5} (\bibinfo{year}{2014}),
  \bibinfo{pages}{612--624}.
\newblock


\bibitem[\protect\citeauthoryear{Danielle, Alexander, and Nitschke}{Danielle
  et~al\mbox{.}}{2019}]%
        {nagar2019cost}
\bibfield{author}{\bibinfo{person}{Nagar Danielle}, \bibinfo{person}{Furman
  Alexander}, {and} \bibinfo{person}{Geoff. Nitschke}.}
  \bibinfo{year}{2019}\natexlab{}.
\newblock \showarticletitle{The Cost of Big Brains in Groups}. In
  \bibinfo{booktitle}{\emph{Proceedings of the 2019 Conference on Artificial
  Life}}. \bibinfo{publisher}{MIT Press}, \bibinfo{address}{Newcastle, United
  Kingdom}, \bibinfo{pages}{404--411}.
\newblock


\bibitem[\protect\citeauthoryear{Dunbar and Shultz}{Dunbar and Shultz}{2017}]%
        {dunbar2017there}
\bibfield{author}{\bibinfo{person}{Robin~IM Dunbar} {and}
  \bibinfo{person}{Susanne Shultz}.} \bibinfo{year}{2017}\natexlab{}.
\newblock \showarticletitle{Why are there so many explanations for primate
  brain evolution?}
\newblock \bibinfo{journal}{\emph{Philosophical Transactions of the Royal
  Society B: Biological Sciences}} \bibinfo{volume}{372},
  \bibinfo{number}{1727} (\bibinfo{year}{2017}), \bibinfo{pages}{20160244}.
\newblock


\bibitem[\protect\citeauthoryear{Dunn}{Dunn}{1964}]%
        {dunn1964multiple}
\bibfield{author}{\bibinfo{person}{Olive~Jean Dunn}.}
  \bibinfo{year}{1964}\natexlab{}.
\newblock \showarticletitle{Multiple comparisons using rank sums}.
\newblock \bibinfo{journal}{\emph{Technometrics}} \bibinfo{volume}{6},
  \bibinfo{number}{3} (\bibinfo{year}{1964}), \bibinfo{pages}{241--252}.
\newblock


\bibitem[\protect\citeauthoryear{Fristoe, Iwaniuk, and Botero}{Fristoe
  et~al\mbox{.}}{2017}]%
        {fristoe2017big}
\bibfield{author}{\bibinfo{person}{Trevor~S Fristoe}, \bibinfo{person}{Andrew~N
  Iwaniuk}, {and} \bibinfo{person}{Carlos~A Botero}.}
  \bibinfo{year}{2017}\natexlab{}.
\newblock \showarticletitle{Big brains stabilize populations and facilitate
  colonization of variable habitats in birds}.
\newblock \bibinfo{journal}{\emph{Nature ecology \& evolution}}
  \bibinfo{volume}{1}, \bibinfo{number}{11} (\bibinfo{year}{2017}),
  \bibinfo{pages}{1706--1715}.
\newblock


\bibitem[\protect\citeauthoryear{Gerstner, Kistler, Naud, and
  Paninski}{Gerstner et~al\mbox{.}}{2014}]%
        {gerstner2014neuronal}
\bibfield{author}{\bibinfo{person}{Wulfram Gerstner}, \bibinfo{person}{Werner~M
  Kistler}, \bibinfo{person}{Richard Naud}, {and} \bibinfo{person}{Liam
  Paninski}.} \bibinfo{year}{2014}\natexlab{}.
\newblock \bibinfo{booktitle}{\emph{Neuronal dynamics: From single neurons to
  networks and models of cognition}}.
\newblock \bibinfo{publisher}{Cambridge University Press}.
\newblock


\bibitem[\protect\citeauthoryear{Gould and Vrba}{Gould and Vrba}{1982}]%
        {gould1982exaptation}
\bibfield{author}{\bibinfo{person}{Stephen~Jay Gould} {and}
  \bibinfo{person}{Elisabeth~S Vrba}.} \bibinfo{year}{1982}\natexlab{}.
\newblock \showarticletitle{Exaptation—a missing term in the science of
  form}.
\newblock \bibinfo{journal}{\emph{Paleobiology}} \bibinfo{volume}{8},
  \bibinfo{number}{1} (\bibinfo{year}{1982}), \bibinfo{pages}{4--15}.
\newblock


\bibitem[\protect\citeauthoryear{Graber}{Graber}{2017}]%
        {graber2017social}
\bibfield{author}{\bibinfo{person}{Maria~Sereina Graber}.}
  \bibinfo{year}{2017}\natexlab{}.
\newblock \emph{\bibinfo{title}{Social and ecological aspects of brain size
  evolution: a comparative approach}}.
\newblock \bibinfo{thesistype}{Ph.\,D. Dissertation}.
  \bibinfo{school}{University of Zurich}.
\newblock


\bibitem[\protect\citeauthoryear{Heesom-Green, Shock, and
  Nitschke}{Heesom-Green et~al\mbox{.}}{2025}]%
        {heesom2025energy}
\bibfield{author}{\bibinfo{person}{Sian Heesom-Green},
  \bibinfo{person}{Jonathan Shock}, {and} \bibinfo{person}{Geoff Nitschke}.}
  \bibinfo{year}{2025}\natexlab{}.
\newblock \showarticletitle{Energy Costs and Neural Complexity Evolution in
  Changing Environments}. In \bibinfo{booktitle}{\emph{Artificial Life
  Conference Proceedings 37}}, Vol.~\bibinfo{volume}{2025}. MIT Press One
  Rogers Street, Cambridge, MA 02142-1209, USA journals-info~…,
  \bibinfo{pages}{60}.
\newblock


\bibitem[\protect\citeauthoryear{Heldstab, Isler, Graber, Schuppli, and van
  Schaik}{Heldstab et~al\mbox{.}}{2022}]%
        {heldstab2022economics}
\bibfield{author}{\bibinfo{person}{Sandra~A Heldstab}, \bibinfo{person}{Karin
  Isler}, \bibinfo{person}{Sereina~M Graber}, \bibinfo{person}{Caroline
  Schuppli}, {and} \bibinfo{person}{Carel~P van Schaik}.}
  \bibinfo{year}{2022}\natexlab{}.
\newblock \showarticletitle{The economics of brain size evolution in
  vertebrates}.
\newblock \bibinfo{journal}{\emph{Current Biology}} \bibinfo{volume}{32},
  \bibinfo{number}{12} (\bibinfo{year}{2022}), \bibinfo{pages}{R697--R708}.
\newblock


\bibitem[\protect\citeauthoryear{Hirsch, Kays, Alavi, Caillaud, Havmoller,
  Mares, and Crofoot}{Hirsch et~al\mbox{.}}{2024}]%
        {hirsch2024smarter}
\bibfield{author}{\bibinfo{person}{Ben~T Hirsch}, \bibinfo{person}{Roland
  Kays}, \bibinfo{person}{Shauhin Alavi}, \bibinfo{person}{Damien Caillaud},
  \bibinfo{person}{Rasmus Havmoller}, \bibinfo{person}{Rafael Mares}, {and}
  \bibinfo{person}{Margaret Crofoot}.} \bibinfo{year}{2024}\natexlab{}.
\newblock \showarticletitle{Smarter foragers do not forage smarter: a test of
  the diet hypothesis for brain expansion}.
\newblock \bibinfo{journal}{\emph{Proceedings of the Royal Society B}}
  \bibinfo{volume}{291}, \bibinfo{number}{2023} (\bibinfo{year}{2024}),
  \bibinfo{pages}{20240138}.
\newblock


\bibitem[\protect\citeauthoryear{Isler and van Schaik}{Isler and van
  Schaik}{2009}]%
        {isler2009expensive}
\bibfield{author}{\bibinfo{person}{Karin Isler} {and} \bibinfo{person}{Carel~P
  van Schaik}.} \bibinfo{year}{2009}\natexlab{}.
\newblock \showarticletitle{The expensive brain: a framework for explaining
  evolutionary changes in brain size}.
\newblock \bibinfo{journal}{\emph{Journal of human evolution}}
  \bibinfo{volume}{57}, \bibinfo{number}{4} (\bibinfo{year}{2009}),
  \bibinfo{pages}{392--400}.
\newblock


\bibitem[\protect\citeauthoryear{Kruskal and Wallis}{Kruskal and
  Wallis}{1952}]%
        {kruskal1952use}
\bibfield{author}{\bibinfo{person}{William~H Kruskal} {and}
  \bibinfo{person}{W~Allen Wallis}.} \bibinfo{year}{1952}\natexlab{}.
\newblock \showarticletitle{Use of ranks in one-criterion variance analysis}.
\newblock \bibinfo{journal}{\emph{Journal of the American statistical
  Association}} \bibinfo{volume}{47}, \bibinfo{number}{260}
  (\bibinfo{year}{1952}), \bibinfo{pages}{583--621}.
\newblock


\bibitem[\protect\citeauthoryear{Luo, Zhong, Huang, Li, Liao, and
  Kotrschal}{Luo et~al\mbox{.}}{2017}]%
        {luo2017seasonality}
\bibfield{author}{\bibinfo{person}{Yi Luo}, \bibinfo{person}{Mao~Jun Zhong},
  \bibinfo{person}{Yan Huang}, \bibinfo{person}{Feng Li},
  \bibinfo{person}{Wen~Bo Liao}, {and} \bibinfo{person}{Alexander Kotrschal}.}
  \bibinfo{year}{2017}\natexlab{}.
\newblock \showarticletitle{Seasonality and brain size are negatively
  associated in frogs: evidence for the expensive brain framework}.
\newblock \bibinfo{journal}{\emph{Scientific reports}} \bibinfo{volume}{7},
  \bibinfo{number}{1} (\bibinfo{year}{2017}), \bibinfo{pages}{16629}.
\newblock


\bibitem[\protect\citeauthoryear{Mailer, Nitschke, and Raw}{Mailer
  et~al\mbox{.}}{2021}]%
        {mailer2021evolving}
\bibfield{author}{\bibinfo{person}{Chris Mailer}, \bibinfo{person}{Geoff
  Nitschke}, {and} \bibinfo{person}{Leanne Raw}.}
  \bibinfo{year}{2021}\natexlab{}.
\newblock \showarticletitle{Evolving gaits for damage control in a hexapod
  robot}. In \bibinfo{booktitle}{\emph{Proceedings of the Genetic and
  Evolutionary Computation Conference}}. \bibinfo{pages}{146--153}.
\newblock


\bibitem[\protect\citeauthoryear{Mettke-Hofmann}{Mettke-Hofmann}{2014}]%
        {mettke2014cognitive}
\bibfield{author}{\bibinfo{person}{Claudia Mettke-Hofmann}.}
  \bibinfo{year}{2014}\natexlab{}.
\newblock \showarticletitle{Cognitive ecology: ecological factors, life-styles,
  and cognition}.
\newblock \bibinfo{journal}{\emph{Wiley Interdisciplinary Reviews: Cognitive
  Science}} \bibinfo{volume}{5}, \bibinfo{number}{3} (\bibinfo{year}{2014}),
  \bibinfo{pages}{345--360}.
\newblock


\bibitem[\protect\citeauthoryear{Michaud, Toussaint, and Gilissen}{Michaud
  et~al\mbox{.}}{2022}]%
        {michaud2022impact}
\bibfield{author}{\bibinfo{person}{Margot Michaud}, \bibinfo{person}{SLD
  Toussaint}, {and} \bibinfo{person}{Emmanuel Gilissen}.}
  \bibinfo{year}{2022}\natexlab{}.
\newblock \showarticletitle{The impact of environmental factors on the
  evolution of brain size in carnivorans}.
\newblock \bibinfo{journal}{\emph{Communications Biology}} \bibinfo{volume}{5},
  \bibinfo{number}{1} (\bibinfo{year}{2022}), \bibinfo{pages}{998}.
\newblock


\bibitem[\protect\citeauthoryear{Miikkulainen}{Miikkulainen}{2025}]%
        {miikkulainen2025neuroevolution}
\bibfield{author}{\bibinfo{person}{Risto Miikkulainen}.}
  \bibinfo{year}{2025}\natexlab{}.
\newblock \showarticletitle{Neuroevolution insights into biological neural
  computation}.
\newblock \bibinfo{journal}{\emph{Science}} \bibinfo{volume}{387},
  \bibinfo{number}{6735} (\bibinfo{year}{2025}), \bibinfo{pages}{eadp7478}.
\newblock


\bibitem[\protect\citeauthoryear{Nagar, Furman, and Nitschke}{Nagar
  et~al\mbox{.}}{2019}]%
        {nagar2019costBodies}
\bibfield{author}{\bibinfo{person}{Danielle Nagar}, \bibinfo{person}{Alexander
  Furman}, {and} \bibinfo{person}{Geoff Nitschke}.}
  \bibinfo{year}{2019}\natexlab{}.
\newblock \showarticletitle{The cost of complexity in robot bodies}. In
  \bibinfo{booktitle}{\emph{2019 IEEE Congress on Evolutionary Computation
  (CEC)}}. IEEE, \bibinfo{pages}{2713--2720}.
\newblock


\bibitem[\protect\citeauthoryear{Pfeiffer, Nessler, Douglas, and
  Maass}{Pfeiffer et~al\mbox{.}}{2010}]%
        {pfeiffer2010reward}
\bibfield{author}{\bibinfo{person}{Michael Pfeiffer}, \bibinfo{person}{Bernhard
  Nessler}, \bibinfo{person}{Rodney~J Douglas}, {and} \bibinfo{person}{Wolfgang
  Maass}.} \bibinfo{year}{2010}\natexlab{}.
\newblock \showarticletitle{Reward-modulated Hebbian learning of decision
  making}.
\newblock \bibinfo{journal}{\emph{Neural computation}} \bibinfo{volume}{22},
  \bibinfo{number}{6} (\bibinfo{year}{2010}), \bibinfo{pages}{1399--1444}.
\newblock


\bibitem[\protect\citeauthoryear{Sayol, Maspons, Lapiedra, Iwaniuk,
  Sz{\'e}kely, and Sol}{Sayol et~al\mbox{.}}{2016}]%
        {sayol2016environmental}
\bibfield{author}{\bibinfo{person}{Ferran Sayol}, \bibinfo{person}{Joan
  Maspons}, \bibinfo{person}{Oriol Lapiedra}, \bibinfo{person}{Andrew~N
  Iwaniuk}, \bibinfo{person}{Tam{\'a}s Sz{\'e}kely}, {and}
  \bibinfo{person}{Daniel Sol}.} \bibinfo{year}{2016}\natexlab{}.
\newblock \showarticletitle{Environmental variation and the evolution of large
  brains in birds}.
\newblock \bibinfo{journal}{\emph{Nature communications}} \bibinfo{volume}{7},
  \bibinfo{number}{1} (\bibinfo{year}{2016}), \bibinfo{pages}{13971}.
\newblock


\bibitem[\protect\citeauthoryear{Schuck-Paim, Alonso, and Ottoni}{Schuck-Paim
  et~al\mbox{.}}{2008}]%
        {schuck2008cognition}
\bibfield{author}{\bibinfo{person}{Cynthia Schuck-Paim},
  \bibinfo{person}{Wladimir~J Alonso}, {and} \bibinfo{person}{Eduardo~B
  Ottoni}.} \bibinfo{year}{2008}\natexlab{}.
\newblock \showarticletitle{Cognition in an ever-changing world: climatic
  variability is associated with brain size in neotropical parrots}.
\newblock \bibinfo{journal}{\emph{Brain, Behavior and Evolution}}
  \bibinfo{volume}{71}, \bibinfo{number}{3} (\bibinfo{year}{2008}),
  \bibinfo{pages}{200--215}.
\newblock


\bibitem[\protect\citeauthoryear{Sol}{Sol}{2009}]%
        {sol2009revisiting}
\bibfield{author}{\bibinfo{person}{Daniel Sol}.}
  \bibinfo{year}{2009}\natexlab{}.
\newblock \showarticletitle{Revisiting the cognitive buffer hypothesis for the
  evolution of large brains}.
\newblock \bibinfo{journal}{\emph{Biology letters}} \bibinfo{volume}{5},
  \bibinfo{number}{1} (\bibinfo{year}{2009}), \bibinfo{pages}{130--133}.
\newblock


\bibitem[\protect\citeauthoryear{Sol, Bacher, Reader, and Lefebvre}{Sol
  et~al\mbox{.}}{2008}]%
        {sol2008brain}
\bibfield{author}{\bibinfo{person}{Daniel Sol}, \bibinfo{person}{Sven Bacher},
  \bibinfo{person}{Simon~M Reader}, {and} \bibinfo{person}{Louis Lefebvre}.}
  \bibinfo{year}{2008}\natexlab{}.
\newblock \showarticletitle{Brain size predicts the success of mammal species
  introduced into novel environments}.
\newblock \bibinfo{journal}{\emph{the american naturalist}}
  \bibinfo{volume}{172}, \bibinfo{number}{S1} (\bibinfo{year}{2008}),
  \bibinfo{pages}{S63--S71}.
\newblock


\bibitem[\protect\citeauthoryear{Sol, Duncan, Blackburn, Cassey, and
  Lefebvre}{Sol et~al\mbox{.}}{2005}]%
        {sol2005big}
\bibfield{author}{\bibinfo{person}{Daniel Sol}, \bibinfo{person}{Richard~P
  Duncan}, \bibinfo{person}{Tim~M Blackburn}, \bibinfo{person}{Phillip Cassey},
  {and} \bibinfo{person}{Louis Lefebvre}.} \bibinfo{year}{2005}\natexlab{}.
\newblock \showarticletitle{Big brains, enhanced cognition, and response of
  birds to novel environments}.
\newblock \bibinfo{journal}{\emph{Proceedings of the National Academy of
  Sciences}} \bibinfo{volume}{102}, \bibinfo{number}{15}
  (\bibinfo{year}{2005}), \bibinfo{pages}{5460--5465}.
\newblock


\bibitem[\protect\citeauthoryear{Sol and Lefebvre}{Sol and Lefebvre}{2000}]%
        {sol2000behavioural}
\bibfield{author}{\bibinfo{person}{Daniel Sol} {and} \bibinfo{person}{Louis
  Lefebvre}.} \bibinfo{year}{2000}\natexlab{}.
\newblock \showarticletitle{Behavioural flexibility predicts invasion success
  in birds introduced to New Zealand}.
\newblock \bibinfo{journal}{\emph{Oikos}} \bibinfo{volume}{90},
  \bibinfo{number}{3} (\bibinfo{year}{2000}), \bibinfo{pages}{599--605}.
\newblock


\bibitem[\protect\citeauthoryear{Stanley and Miikkulainen}{Stanley and
  Miikkulainen}{2002}]%
        {stanley2002evolving}
\bibfield{author}{\bibinfo{person}{Kenneth Stanley} {and}
  \bibinfo{person}{Risto Miikkulainen}.} \bibinfo{year}{2002}\natexlab{}.
\newblock \showarticletitle{Evolving neural networks through augmenting
  topologies}.
\newblock \bibinfo{journal}{\emph{Evolutionary computation}}
  \bibinfo{volume}{10}, \bibinfo{number}{2} (\bibinfo{year}{2002}),
  \bibinfo{pages}{99--127}.
\newblock


\bibitem[\protect\citeauthoryear{van Woerden}{van Woerden}{2011}]%
        {van2011influence}
\bibfield{author}{\bibinfo{person}{Janneke~T van Woerden}.}
  \bibinfo{year}{2011}\natexlab{}.
\newblock \emph{\bibinfo{title}{The influence of seasonality on brain size
  evolution in primates}}.
\newblock \bibinfo{thesistype}{Ph.\,D. Dissertation}.
  \bibinfo{school}{University of Zurich}.
\newblock


\bibitem[\protect\citeauthoryear{Van~Woerden, Van~Schaik, and
  Isler}{Van~Woerden et~al\mbox{.}}{2010}]%
        {van2010effects}
\bibfield{author}{\bibinfo{person}{Janneke~T Van~Woerden},
  \bibinfo{person}{Carel~P Van~Schaik}, {and} \bibinfo{person}{Karin Isler}.}
  \bibinfo{year}{2010}\natexlab{}.
\newblock \showarticletitle{Effects of seasonality on brain size evolution:
  evidence from strepsirrhine primates}.
\newblock \bibinfo{journal}{\emph{The American Naturalist}}
  \bibinfo{volume}{176}, \bibinfo{number}{6} (\bibinfo{year}{2010}),
  \bibinfo{pages}{758--767}.
\newblock


\bibitem[\protect\citeauthoryear{Van~Woerden, Willems, van Schaik, and
  Isler}{Van~Woerden et~al\mbox{.}}{2012}]%
        {van2012large}
\bibfield{author}{\bibinfo{person}{Janneke~T Van~Woerden},
  \bibinfo{person}{Erik~P Willems}, \bibinfo{person}{Carel~P van Schaik}, {and}
  \bibinfo{person}{Karin Isler}.} \bibinfo{year}{2012}\natexlab{}.
\newblock \showarticletitle{Large brains buffer energetic effects of seasonal
  habitats in catarrhine primates}.
\newblock \bibinfo{journal}{\emph{Evolution}} \bibinfo{volume}{66},
  \bibinfo{number}{1} (\bibinfo{year}{2012}), \bibinfo{pages}{191--199}.
\newblock


\bibitem[\protect\citeauthoryear{Vincze}{Vincze}{2016}]%
        {vincze2016light}
\bibfield{author}{\bibinfo{person}{Orsolya Vincze}.}
  \bibinfo{year}{2016}\natexlab{}.
\newblock \showarticletitle{Light enough to travel or wise enough to stay?
  Brain size evolution and migratory behavior in birds}.
\newblock \bibinfo{journal}{\emph{Evolution}} \bibinfo{volume}{70},
  \bibinfo{number}{9} (\bibinfo{year}{2016}), \bibinfo{pages}{2123--2133}.
\newblock


\bibitem[\protect\citeauthoryear{Wagnon and Brown}{Wagnon and Brown}{2020}]%
        {wagnon2020smaller}
\bibfield{author}{\bibinfo{person}{Gigi~S Wagnon} {and}
  \bibinfo{person}{Charles~R Brown}.} \bibinfo{year}{2020}\natexlab{}.
\newblock \showarticletitle{Smaller brained cliff swallows are more likely to
  die during harsh weather}.
\newblock \bibinfo{journal}{\emph{Biology Letters}} \bibinfo{volume}{16},
  \bibinfo{number}{7} (\bibinfo{year}{2020}), \bibinfo{pages}{20200264}.
\newblock


\bibitem[\protect\citeauthoryear{Watson and Nitschke}{Watson and
  Nitschke}{2015}]%
        {WatsonNitschke2015}
\bibfield{author}{\bibinfo{person}{James Watson} {and} \bibinfo{person}{Geoff
  Nitschke}.} \bibinfo{year}{2015}\natexlab{}.
\newblock \showarticletitle{{Evolving Robust Robot Team Morphologies for
  Collective Construction}}. In \bibinfo{booktitle}{\emph{Proceedings of the
  IEEE Symposium Series on Computational Intelligence}}. IEEE Press,
  \bibinfo{address}{Cape Town, South Africa}, \bibinfo{pages}{1039--1046}.
\newblock


\bibitem[\protect\citeauthoryear{Weisbecker, Blomberg, Goldizen, Brown, and
  Fisher}{Weisbecker et~al\mbox{.}}{2015}]%
        {weisbecker2015evolution}
\bibfield{author}{\bibinfo{person}{Vera Weisbecker}, \bibinfo{person}{Simon
  Blomberg}, \bibinfo{person}{Anne~W Goldizen}, \bibinfo{person}{Meredeth
  Brown}, {and} \bibinfo{person}{Diana Fisher}.}
  \bibinfo{year}{2015}\natexlab{}.
\newblock \showarticletitle{The evolution of relative brain size in marsupials
  is energetically constrained but not driven by behavioral complexity}.
\newblock \bibinfo{journal}{\emph{Brain, behavior and evolution}}
  \bibinfo{volume}{85}, \bibinfo{number}{2} (\bibinfo{year}{2015}),
  \bibinfo{pages}{125--135}.
\newblock


\end{thebibliography}







\end{document}